\documentclass[runningheads]{llncs}

\usepackage{eccv}

\usepackage{eccvabbrv}

\usepackage{floatrow}
\newfloatcommand{capbtabbox}{table}[][\FBwidth]
\usepackage{graphicx}
\usepackage{booktabs}
\usepackage{tabularx}
\usepackage{multirow}
\usepackage{amssymb} 
\usepackage{bbding}  
\usepackage{soul}  
\usepackage{colortbl} 
\usepackage{algorithm}  
\usepackage{algpseudocode}  
\usepackage{amsmath}

\usepackage{xcolor}
\usepackage{soul}

\usepackage[accsupp]{axessibility}  

\usepackage[pagebackref,breaklinks,colorlinks]{hyperref}

\usepackage{orcidlink}

\begin{document}

\title{Content-Aware Radiance Fields: Aligning Model Complexity with Scene Intricacy Through Learned Bitwidth Quantization}

\titlerunning{Content-Aware Radiance Fields}

\author{Weihang Liu\inst{1,*}\orcidlink{0009-0009-5760-6437} \and
Xue Xian Zheng \inst{2,*}\orcidlink{0000-0003-4546-8531} \and
Jingyi Yu\inst{1,3}\orcidlink{0000-0001-7196-9861}\and
Xin Lou\inst{1,3,\dag}\orcidlink{0000-0002-8580-0036}
}

\authorrunning{W.~Liu, X.~X.~Zheng et al.}

\institute{ShanghaiTech University 
\and
King Abdullah University of Science and Technology
\\ \and
Key Laboratory of Intelligent Perception and Human-Machine Collaboration
}



\maketitle
\def\thefootnote{*}\footnotetext{Equal Contribution.}
\def\thefootnote{\dag}\footnotetext{Corresponding author.}

\begin{abstract}
The recent popular radiance field models, exemplified by Neural Radiance Fields (NeRF), Instant-NGP and 3D Gaussian Splatting, are designed to represent 3D content by that training models for each individual scene. This unique characteristic of scene representation and per-scene training distinguishes radiance field models from other neural models, because complex scenes necessitate models with higher representational capacity and vice versa.
In this paper, we propose content-aware radiance fields, aligning the model complexity with the scene intricacies through Adversarial Content-Aware Quantization (A-CAQ). 
Specifically, we make the bitwidth of parameters differentiable and trainable, tailored to the unique characteristics of specific scenes and requirements.
The proposed framework has been assessed on Instant-NGP, a well-known NeRF variant and evaluated using various datasets.
Experimental results demonstrate a notable reduction in computational complexity, while preserving the requisite reconstruction and rendering quality, making it beneficial for practical deployment of radiance fields models. 
Codes are available at 
\url{https://github.com/WeihangLiu2024/Content_Aware_NeRF}.




\end{abstract}

\section{Introduction}
\label{sec:intro}




Triggered by the phenomenal Neural Radiance Fields (NeRF)\cite{NeRF}, the idea of representing 3D scenes using trainable models has been widely adopted in many reconstruction and rendering applications. Different from traditional explicit representations like meshes and point clouds, radiance field techniques encode a 3D scene with learnable parameters, which are obtained by training models based on sparse samples of the scene. Despite its advantages, the high-quality representation and rendering offered by radiance fields come at the expense of significant computational complexity. This challenge has been extensively studied, leading to numerous research endeavors aimed at exploring more computational efficient radiance field models \cite{FasrNeRF, SqueezeNeRF, luo2021convolutional,KiloNeRF}.

\begin{figure}[tb]
  \centering
  \includegraphics[width=1.0\textwidth]{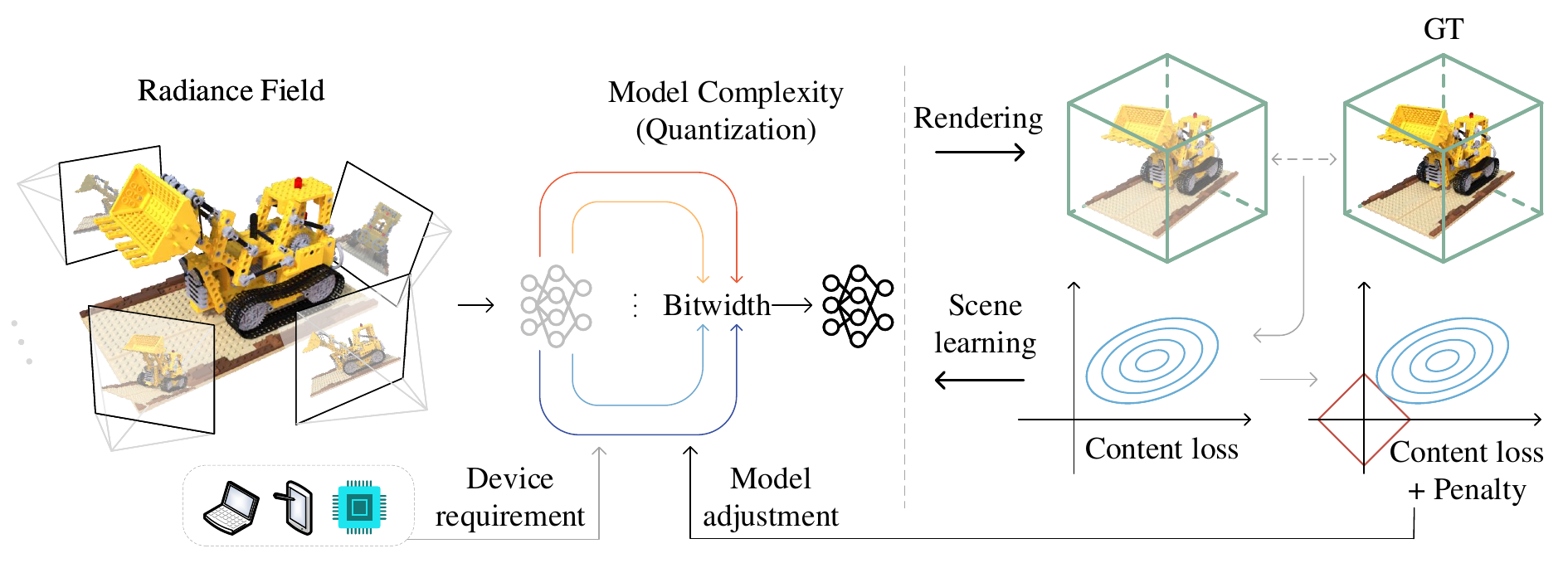}
  \caption{
  \textbf{Overview of content-aware radiance fields.} In this work, model complexity is aligned with scene intricacy through learned bitwidth quantization.} 
  \label{fig:concept}
\end{figure}

While many variants of NeRF have significantly improved efficiency, they typically compress all scenes to a single fixed scale. 
Very few of them consider dealing with the scene contents differently, i.e., employing a \textit{content-aware} strategy to align model complexity with scene intricacy. This principle, as illustrated in \cref{fig:concept}, forms the cornerstone of our work in this paper. 
It is unique to radiance fields due to their distinctive attributes of scene representation and per-scene training.
Intuitively, detailed scenes rich in geometry and texture necessitate more sophisticated radiance field models to capture their nuances. While on the other hand, encoding less complex scenes with simpler models not only suffices for adequate representation but also enhances the efficiency by reducing computational and memory requirements. 


Quantization has been proven a fundamental yet effective technique for reducing model complexity. 
The selection of parameter bitwidth is crucial for balancing computation and performance, bridging the gap between content information loss and computational efficiency.
Recent works have investigated this technique on radiance field models by quantizing them to a pre-defined fixed bitwidth \cite{ICARUS} as well as learning quantization range \cite{MCUNeRF}. 
Nevertheless, these methods often rely on extensive human expertise to select appropriate quantization parameters, resorting to trial and error approaches. Moreover, reducing the bitwidth with controllable performance loss is impractical within their frameworks.


\begin{figure}[tb]
  \centering
  \begin{subfigure}{0.6\linewidth}
    \centerline{\includegraphics[width=\textwidth] 
        {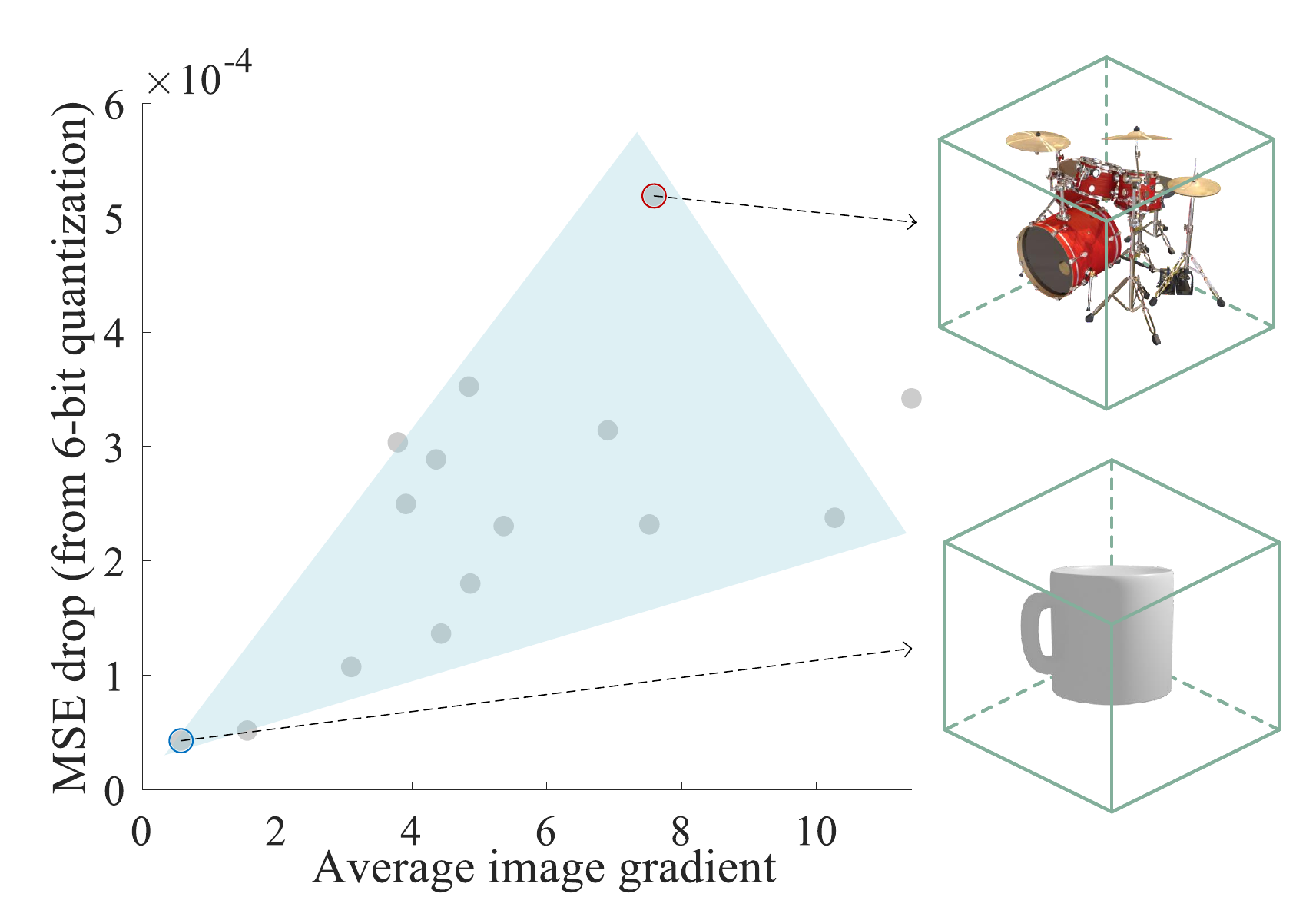} }
    \caption{Quantization sensitivity \wrt average image gradient.}
    \label{fig:intro-a}
  \end{subfigure}
  \hfill
  \begin{subfigure}{0.39\linewidth}
    \centerline{\includegraphics[width=\textwidth] 
        {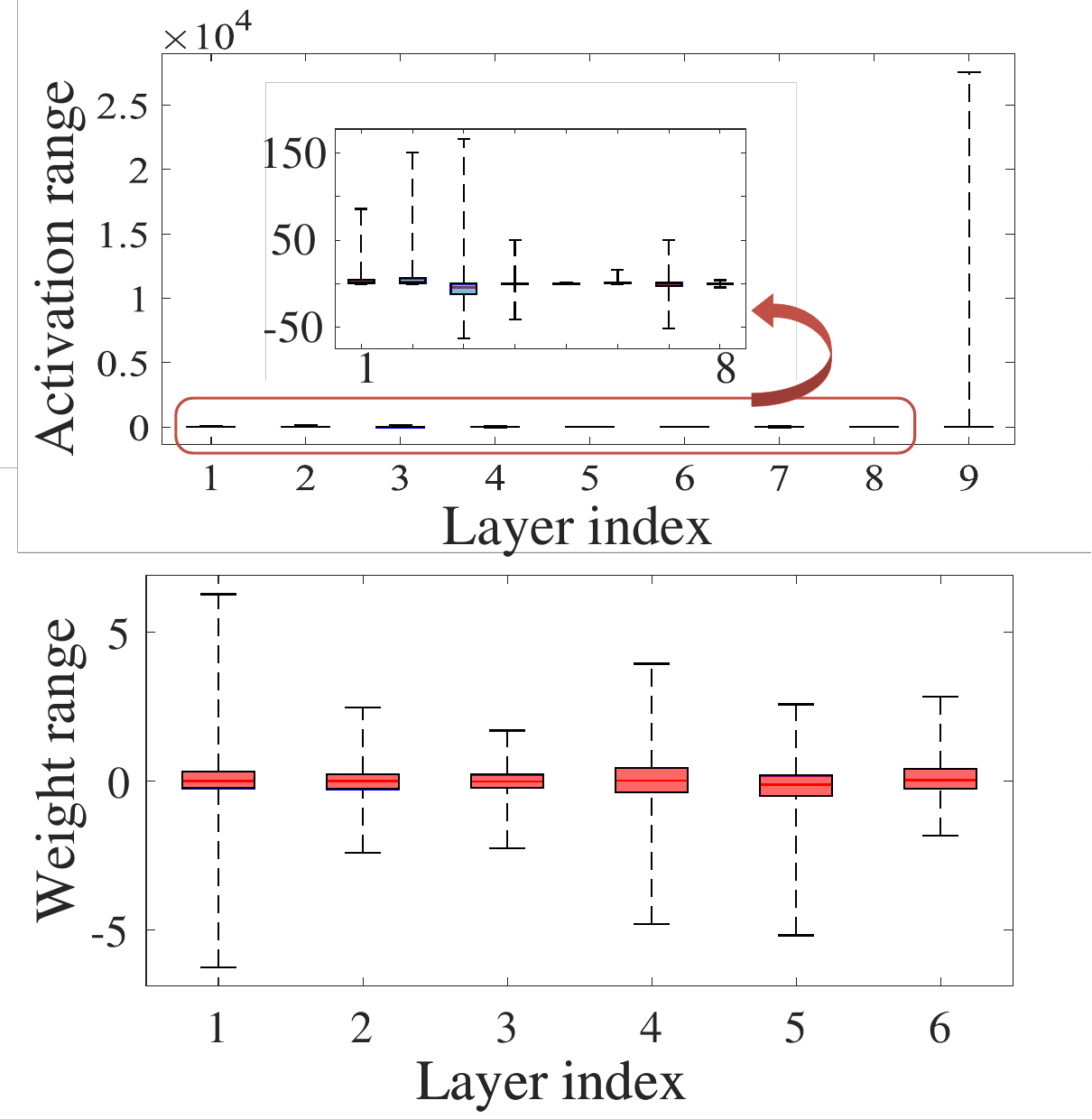} }
    \caption{Variable ranges for components.}
    \label{fig:intro-b}
  \end{subfigure}
  \caption{\textbf{The insights of the proposed LBQ.} 
  (a) The correlation between quantization sensitivity and scene complexity measured using average image gradient of the training set.
  Scenes with complex (simple) geometry and texture suffer more (less) accuracy degradation from quantization.
  (b) Exhibits the notable distinction of variables' distribution among different components.
  Those distributed in large (small) range is required to be quantized with high (low) bitwidth.
  }
  \label{fig:intro}
\end{figure}

\begin{figure}[tb]
    \centering
    \begin{subfigure}{0.32\linewidth}
        \centerline{\includegraphics[width=\textwidth] 
            {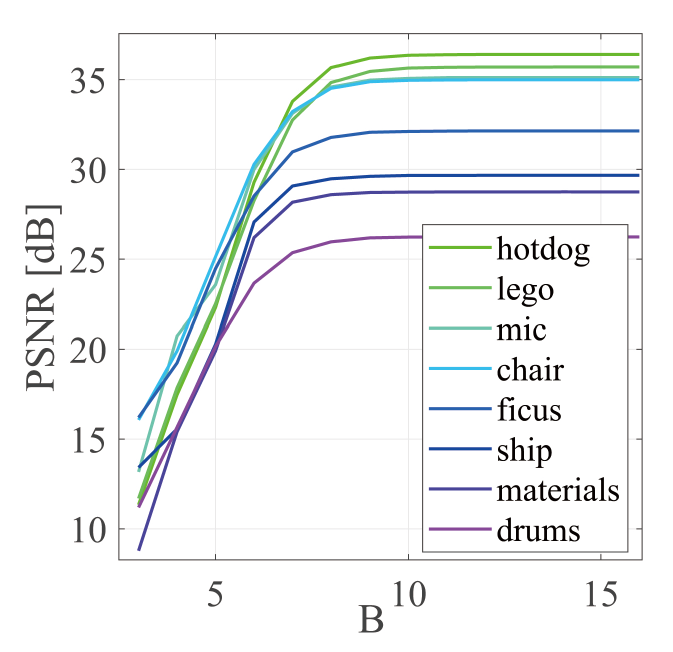} }
        \caption{Scene-wise quantization.}
        \label{fig:quantize_allbit-a}
    \end{subfigure}
    \hfill
    \begin{subfigure}{0.32\linewidth}
        \centerline{\includegraphics[width=\textwidth] 
            {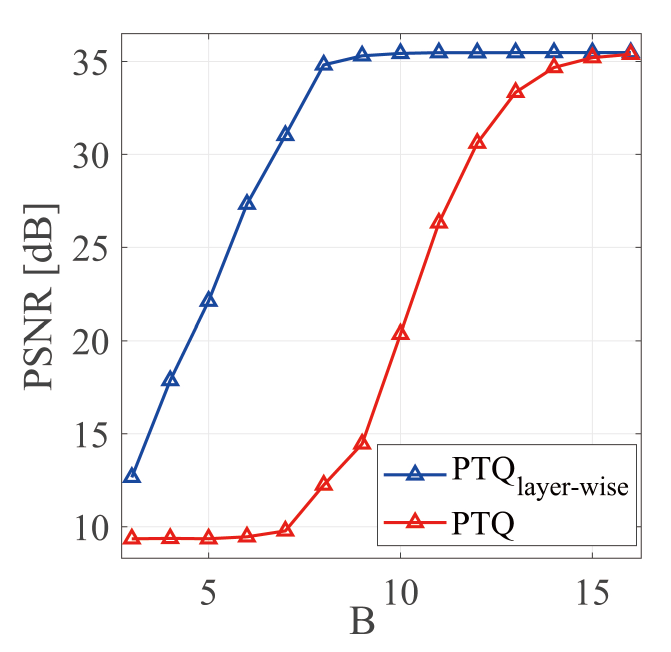} }
        \caption{Layer-wise quantization.}
        \label{fig:quantize_allbit-b}
    \end{subfigure}
    \hfill
    \begin{subfigure}{0.32\linewidth}
        \centerline{\includegraphics[width=\textwidth] 
            {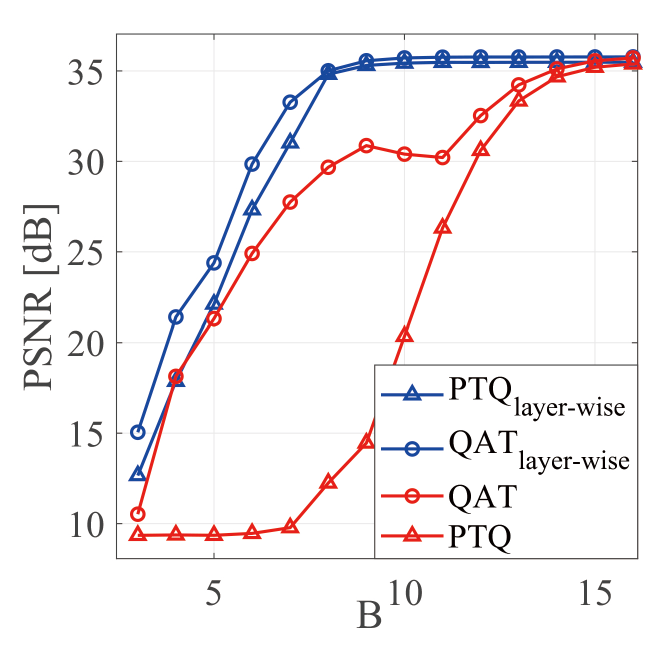} }
        \caption{PTQ v.s. QAT}
        \label{fig:quantize_allbit-c}
    \end{subfigure}
    \caption{\textbf{The insights of the proposed A-CAQ.}
    (a) Layer-wise quantization results for different scenes with various bitwidths, which 
    reveals content-aware characteristics of quantization effects.
    (b) Results of layer-wise and non layer-wise quantization for the "lego" scene. The huge accuracy gap verifies the significance of mixed-precision models.
    (c) QAT alleviates performance degradation as quantization error expands.
  }
  \label{fig:quantize_allbit}
\end{figure}

In this work, we introduce the concept of content-aware radiance fields, which adaptively quantize models by exploiting the content differences among scenes and the features of each layer. The quantization bitwidth of radiance models, which directly correlates with computational complexity, are therefore content-aware, i.e., intricate scenes are accommodated with higher bitwidth models, while simpler scenes utilize lower bitwidth counterparts to reduce computational complexity. 
The key insights for our method are illustrated in \cref{fig:intro}, where average image gradient is used as an estimator for the complexity of the scenes. 
Results in \cref{fig:intro-a} indicate that
scenes with complex structure and texture suffer more from quantization than those with lower complexity. 
Besides, to achieve integer-only inference in rendering, all layers are required to be quantized by considering the significant statistical distinction of output features through the entire pipeline (shown in \cref{fig:intro-b}).
Moreover, the results in \cref{fig:quantize_allbit} demonstrate that scene-wise as well as layer-wise quantization is beneficial and quantization-aware training (QAT) can effectively combat quantization degradation.


Specifically, to establish the connection between quantization sensitivity and bitwidth, we propose Learned Bitwidth Quantization (LBQ) framework illustrated in \cref{fig:pipeline}, facilitating learnable bitwidth based on reconstructed contents.
To fully extract the representation capability of quantized models, Adversarial Content-Aware Quantization (A-CAQ) algorithm is further proposed, which 
searches scene-dependent bitwidth and optimizes radiance fields simultaneously.
The main contributions of this work, as illustrated in \cref{fig:concept}, are as follows:
\begin{itemize}
     \item We introduce content-aware radiance fields that aligns the complexity of models with the intricacies of scenes through quantization, showing that the parameter bitwidth bridges the gap between the content information loss and computational efficiency. 
    \item We introduce the LBQ framework which models the integer bitwidth with "soft bitwidth". This approach makes bitwidth differentiable from the content information loss during training, thereby obviating extensive human expertise to select bitwidths through a trial and error approach. 
    \item We propose A-CAQ that integrates LBQ to penalize layer-wise bitwidth, enabling dynamic learning of lower bitwidth with negligible reconstruction and rendering quality loss. Experimental results on various datasets confirm the superiority of the proposed method.
\end{itemize}

\section{Related Work and Motivation}

\subsubsection{Scene representations and radiance fields.}
Various follow-up works on NeRF have focused specifically on improving computational efficiency.
Rendering based on simplified representations is much more efficient while maintaining quality, which motivates studies of compression for scene primitives.
A number of recent works achieve this goal from different perspectives \cite{nerf_compression1, nerf_compression2, NGP, VQ-AD, Plenoxel, TensoRF, SHACIRA}. 
Specifically, by employing a learnable codebook along with a compact MLP \cite{NGP, VQ-AD}, Instant-NGP mitigates representation workload of network parameters, resulting in superior results in terms of both rendering quality and speed among NeRF variants.


While content information is easily acquired by explicit primitives, it is challenging for implicit representations. 
VQ-AD \cite{VQ-AD} and SHACIRA \cite{SHACIRA} introduced scalable compression of trainable feature grid which render content with different qualities. 
Recursive NeRF \cite{Recursive_NeRF} proposes to dynamically grow the network, considering complexity variations of different patches within one scene. 
However, none of these works are metric-oriented or consider the complexity of different scenes. 
The search for the most suitable compression, considering both content and available resources, remains an empirical task.



\subsubsection{Model quantization.}
Network quantization has been demonstrated as a potent technique for compressing neural models \cite{polino2018model, coushu1, coushu2, coushu3,coushu4}. 
Generally, Post-Training Quantization (PTQ)\cite{MMSE_PTQ,PTQ_recons} and 
Quantization-Aware Training (QAT)\cite{MEBQAT} are two common methods for model quantization. 
Some recent works endeavor to dynamically allocate bitwidths based on the features of each layer, resulting in mixed-precision quantization \cite{mix-precision1, mix-precision2, mix-precision3}.
These mixed-precision methods have succeed in obtaining efficient network for various applications \cite{CADyQ, mix-precision2}.

One significant limitation of existing PTQ and QAT is that they quantize pre-trained models using hyper-defined bitwidths.
Re-quantizing models to other bitwidths can be challenging, as it suffer from differences in statistical characteristics of weights and activations \cite{AdaBits}.
One straightforward solution is to find the optimal bitwidth by considering specific application requirements.
CADyQ \cite{CADyQ} employ this idea to quantize super-resolution (SR) image network based on image gradient and standard deviation of features.
Nevertheless, different from SR net based on convolutional neural networks (CNNs), there is no explicit pattern for MLPs features or 3D scene complexity measurement for neural fields.
Therefore, we propose a novel framework that enables bitwidth differentiation, leveraging content-related Mean Squared Error (MSE) to adeptly select optimal scene-dependent layer-wise bitwidth.

\begin{figure}[tb]
  \centering
  \includegraphics[width=1.0\textwidth]{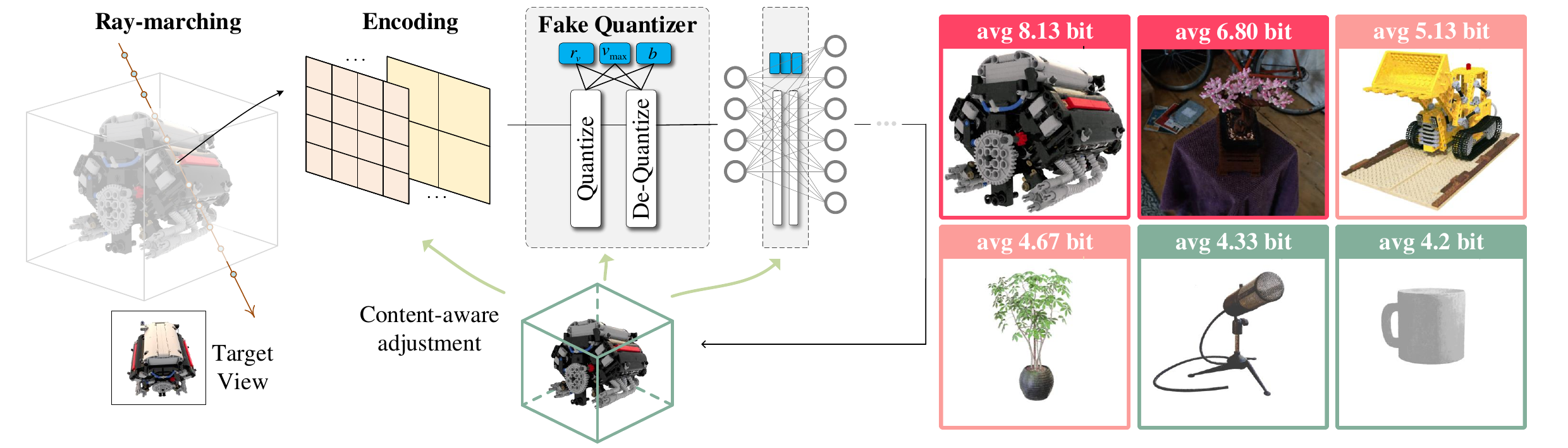}
  \caption{
    \textbf{Overview of the LBQ quantization framework.} 
  The fake quantizers are inserted into different components including encoding and MLPs, which are parameterized with variable range scale, upper bound and bitwidth.
  Quantization error is simulated by quantize and de-quantize procedure and quantization parameters are updated with direction indicated by gradient descent.
  Examples with different complexity are given on the right. 
  }
  \label{fig:pipeline}
\end{figure}

\section{Proposed Method}

\subsection{Preliminary}
\label{sec:preliminary}
\subsubsection{Radiance fields with learnable codebook.} 
Instant-NGP, a well-known NeRF variants incorporating a learnable multi-resolution hash table, is selected as our baseline model due to its inclusion of prevalent feature modalities in radiance fields, namely spatial levels-of-detail (LOD) feature grid and MLPs. 
Specifically, it approximates the mapping from continuous 5D vectors to opacities $\sigma$ and view-dependent colors $\mathbf{c}$, denoted as 
\begin{equation}
    F: (\mathbf{x}, \mathbf{d}) \xrightarrow{} (\sigma, \mathbf{c}),
\end{equation} 
where $\mathbf{x}$ and $\mathbf{d}$ are 3D location coordinate (of sample points) and 2D view direction respectively. Given a multi-resolution feature table with trainable parameters $\mathbf{\Theta}$, the location coordinates are encoded as $\mathbf{y_x}=\text{enc}(\mathbf{x}, \mathbf{\Theta})$, and the view directions are encoded as $\mathbf{y_d}$ with spherical harmonics \cite{NGP}. The encoded vector is then fed to the MLP to produce the color and opacity as
\begin{equation}
    (\sigma, \mathbf{c}) = m(\mathbf{y},\mathbf{\Phi}), 
\end{equation}
where $\mathbf{y} = (\mathbf{y_x}, \mathbf{y_d})$, $\mathbf{\Phi}$ denotes the weights of MLPs.
With the colors and opacities of all sample points, the expected color along a ray $\mathbf{r}$ can be calculated using volume rendering as
\begin{equation}
    \mathbf{C}(\mathbf{r}) = \int^{t_\text{f}}_{t_\text{n}}T(t)
    \sigma( \mathbf{r}(t) )
    \mathbf{c}(\mathbf{r}(t), \mathbf{d}) dt,\ \text{where} \ 
    T(t)=
    \exp(-\int^t_{t_{\text{n}}} \sigma( \mathbf{r}(s) ) ds ),
\end{equation}
$\mathbf{r}(t)=\mathbf{o} + t\mathbf{d}$ are sampled points along the ray defined by origin $\mathbf{o}$ and direction $\mathbf{d}$.

\subsubsection{Quantization.}
To attain integer-only inference, quantization of both parameters (\ie $\mathbf{\Omega} = \{\mathbf{\Theta}, \mathbf{\Phi} \}$) and activations for each layer is imperative.
As shown in \cref{fig:pipeline}, fake quantizers can be inserted to simulate quantization error in training procedure.
Given the data to be quantized $\mathbf{v}$, along with the step size $s$ and zero-point offset $Z$,
the function of a fake quantizer is to dequantize the quantized data
$\hat{\mathbf{V}}$. This operation can be expressed as 
\begin{equation}
    \label{fakeQ}
    \hat{\mathbf{v}} 
    = s(\hat{\mathbf{V}}-Z)
    = s\left[
    \text{clamp}
        \left(
            \left\lfloor{\frac{\mathbf{v}}{s}} \right\rceil + Z;q_{\text{min}}, q_{\text{max}}
        \right)
        -Z
    \right],
\end{equation}
where $\lfloor {\cdot} \rceil$ is the round to nearest operator, and $\text{clamp}(\cdot)$ is defined as 
\begin{equation}
    \label{clamp}
    \text{clamp}(x;a,c) =
    \left\{
        \begin{aligned}
        &a \quad x<a \\
        &x \quad a\leq x \leq c \\
        &c \quad x>c
        \end{aligned}
    \right. .
\end{equation}
Based on \cref{fakeQ}, quantization training can be conducted to mitigate introduced quantization noise.

\subsection{Differentiable Quantization for Radiance Fields}
\label{sec:quan_NeRF}

\subsubsection{Learned Bitwidth Quantization (LBQ).}
The quantization training model has been established in \cref{sec:preliminary}.
In this section, we initially reveal that the prevalent QAT-based quantization technique is inherently limited to fixed-bitwidth scenarios, followed by the introduction of the innovative LBQ scheme used in our content-aware quantization framework.


Existing QAT schemes, such as LSQ \cite{LSQ}, LSQ+ \cite{LSQ+}, aim to directly train the step size and zero-point offset as
\begin{equation}
    s = \frac{v_{\text{max}} - v_{\text{min}} } {q_{\text{max}}-q_{\text{min}}} = \frac{r_v}{r_q},
\end{equation}
\begin{equation}
    Z = \left\lfloor q_{\text{max}}-\frac{v_{\text{max}}}{s} \right\rceil
      = \left\lfloor q_{\text{max}}-\frac{v_{\text{max}}}{r_v} r_q \right\rceil,
\end{equation}
where $r_v$ and $r_q=2^B-1$ are the range scales of $\mathbf{v}$ and $\hat{\mathbf{V}}$, respectively.
To make the "round" operation differentiable, the Straight-Trough-Estimator (STE) \cite{STE} ${\partial \left\lfloor x \right\rceil}/{\partial x}=1$ is assumed. 
It indicates that $s$ and $Z$ can only be regarded as leaf nodes with fixed bitwidth, where $r_q$ is considered a constant.

To further make bitwidth scalable, 
we introduce floating point "soft bitwidth" $b$, where $B = \left\lfloor b \right\rceil$. Moreover, $r_v$, $v_{\text{max}}$ and $b$ are selected as trainable quantization parameters, where $r_v$ and $v_{\text{max}}$ serve as replacements of step size and offset, respectively.
The partial derivatives of \cref{fakeQ} can then be derived as in \cref{tab:derivatives}.

\begin{table}[tb]
  \fontsize{9}{16}\selectfont
  \caption{Derivatives to key parameters.}
  \label{tab:derivatives}
  \centering
  \begin{tabular}{@{}lccc@{}}
    \toprule 
    Variable range & 
    $\partial \hat{v}/\partial r_v$ & 
    $\partial \hat{v}/\partial b$ & 
    $\partial \hat{v}/\partial v_\text{max}$
    \\
    \midrule
    $v_\text{min}\leq v\leq v_\text{max}$ & 
    \colorbox[RGB]{244,241,222}{$\left( s \cdot \left\lfloor v/s \right\rceil - v \right)$}
    $/r_v$ 
    & \colorbox[RGB]{244,241,222}{$\left(v- s \cdot \left\lfloor v/s \right\rceil\right)$} $\cdot2^B\ln2/r_q $ &
    \colorbox[RGB]{231,239,250}{0}\\
    $v > v_\text{max}$ & $1-v_\text{max}/r_v-Z/r_q$ & \multirow{2}{*}{$\left(v_\text{max}-r_v + sZ\right) \cdot 2^B\ln2/r_q$} 
     & \multirow{2}{*}{\colorbox[RGB]{231,239,250}{1}}\\
    $v < v_\text{min}$ & $-v_\text{max}/r_v-Z/r_q$ & & {} \\
  \bottomrule
  \end{tabular}
\end{table}
The foregoing equations elucidate the operational mechanism of the proposed differentiable quantization:
\begin{itemize}
    \item The parameter $v_\text{max}$ represents the offsets of the quantization range, calibrated solely upon the occurrence of overflow, as explicated in \cref{tab:derivatives}.
    \item Both $r_v$ and $b$ are updated in response to observed quantization errors calculated with the specific term $\left(v- s \cdot \left\lfloor v/s \right\rceil\right)$ in \cref{tab:derivatives}.
\end{itemize}
This proposed method makes bitwidth learnable during the training procedure. 

\subsubsection{Quantization schemes for radiance field models.}

The criterion for selecting quantization parameters is to minimize the error summation generated from rounding and overflow \cite{WhitePpaer}.
Considering statistical properties and inference efficiency, we establish three different quantization schemes for different components within the radiance field pipeline.

\textit{Neural weights}. \cite{LSQ+} indicates that symmetric signed quantization is highly recommended for neural weights which are empirically distributed symmetrically around zero. More importantly, 
quantizing weights in this manner introduces no additional computational overhead during inference. 

\textit{ReLU and exponential activations}. As these functions always produce positive output, we use asymmetric unsigned quantization. Similar settings are found in \cite{LSQ}. 
Based on this configuration, lower bitwidth can be reached.

\textit{Positional encoding (PE) and others}. To deal with components lacking significant statistical features, we introduce $v_{\text{max}}$ to represent trainable offset. \cite{WhitePpaer} proves that this configuration will not introduce any additional computational overhead during inference. The learnable codebook is quantized in this manner. 

The training details about these schemes are provided in \cref{tab:quan_scheme}. Calculations of gradients \wrt three different sets of trainable quantization parameters can be found in \emph{Supplementary materials}.

\begin{table}[tb]
  \caption{Quantization schemes for different components in radiance field pipelines.}
  \setlength{\tabcolsep}{2.5mm}
  \label{tab:quan_scheme}
  \centering
  \begin{tabular}{@{}lcccc@{}}
    \toprule
    Module name          & $v_{\text{max}}$ & $r_v$ & $b$  & $[q_{\text{min}}, q_{\text{max}}]$\\
    \midrule
    Neural weights       & \centering N/A        & trainable & trainable & $[-2^{B-1}, 2^{B-1}-1]$ \\
    ReLU and exponential & \centering N/A         & trainable & trainable & $[0, 2^B-1]$            \\
    PE and others        & \centering trainable   & trainable & trainable & $[0, 2^B-1]$         \\
  \bottomrule
  \end{tabular}
\end{table}

\subsection{Adversarial Content-Aware Quantization (A-CAQ)}
\label{subsec:A-CAQ}

As we have illustrated in \cref{fig:intro}, the impact of quantization varies depending on the contents of scenes.
To leverage this characteristic effectively, it is crucial to extract content-related information from radiance fields.
While it is straightforward for explicit representations such as 3D Gaussian Splatting (3DGS) \cite{3DGS}, it presents a formidable challenge for neural fields, primarily owing to the implicit nature of the patterns encapsulated within MLPs.
Nevertheless, MSE between rendered view and ground truth emerges as a fortuitous revelation, serving as a compelling indicator of the precision achieved in reconstructing the 3D content.
Therefore, the MSE loss assumes a pivotal role for ascertaining content-awareness of radiance field modeling.

Utilizing LBQ proposed in \cref{sec:quan_NeRF}, scene-dependent layer-wise quantization schemes can therefore be learned from MSE.
However, bitwidth cannot adhere to the same objective function that guides other parameters. 
The discrepancy arises from the accuracy degradation inevitable introduced by reducing bitwidth, which operates in adversarial manner compared to optimizing MSE.
To address this problem, we proposed to define bitwidth learning loss as
\begin{equation}
\label{eq:loss_b}
    \mathcal{L}^\text{bit} = 
    \sqrt{|| \mathcal{L}^{\text{NeRF}} - \mathcal{L}^{\text{metric}}||} + \sum_{i \in \mathcal{M}} \epsilon_i B_i,
\end{equation}
where $\mathcal{M}=\{1,2,...,M\}$ is the indexes of layers or components, 
$\mathcal{L}^{\text{NeRF}}$, defined as
\begin{equation}
\label{eq:loss_nerf}
    \mathcal{L}^{\text{NeRF}} = \sum_{\ell \in \mathcal{R}}||\hat{C}(\ell) - C(\ell) ||_{\text{F}}^{2},
\end{equation}
is the MSE loss function used for training radiance field \cite{NeRF}, $\mathcal{R}$ is the set of rays in one batch,
and the term $\mathcal{L}^\text{metric}$ is a hyper-defined metric representing the minimal accuracy requirement. 
By employing various loss metrics, rendering quality can be controlled, transfering redundant accuracy to efficiency. 
Weighted bitwidth penalties are further introduced to remove redundant bitwidth that have minimal impact on the overall loss.
In addition, the weights $\epsilon_i$ offer greater flexibility, enabling assignment according to specific requirements.
For example, higher penalty could be allocated to the bitwidth of the learnable codebook to obtain memory-efficient quantization schemes.

Dynamic bitwidth search can be achieved by solely optimizing \cref{eq:loss_b}. 
However, this approach results in learning bitwidth in a PTQ manner. Directly searching for bitwidth in the QAT space is impractical, as QAT accuracy can only be obtained through trial and error. As illustrated in \cref{fig:quantize_allbit-c}, the performance gap between PTQ and QAT grows when quantizing to lower bitwidth. To mitigate this degradation, we propose A-CAQ, a multi-task learning-based method expressed as
\begin{equation}
\label{eq:A-CAQ}
    \mathcal{L}^\text{A-CAQ} = 
    \min_{\mathcal{Q}} \mathcal{L}^\text{NeRF} + 
    \min_{\mathbf{b}} \mathcal{L}^\text{bit}.
\end{equation}
where $\mathcal{Q}=\{\mathbf{\Omega}, \mathbf{v}_\text{max}, \mathbf{r}_v \}$.
This task can be effectively solved by optimizing \cref{eq:loss_b} and \cref{eq:loss_nerf} alternatively.
Detailed pseudo codes are found in \emph{Supplemenetray Materials}.

The interpretation of \cref{eq:A-CAQ} primarily pertains to adversarial learning: minimizing \cref{eq:loss_b} leads to a lower bitwidth solution as well as higher $\mathcal{L}^\text{NeRF}$. 
And a lower bitwidth provides $\mathcal{L}^\text{NeRF}$ with more potential to alleviate accuracy degradation, thereby creating more search space to achieve a lower bitwidth solution.

\section{Experiments}

Experiments are conducted to demonstrate the effectiveness and versatility of our content-aware quantization framework for radiance fields. We first provide the implementation details, including model and training configurations, in \cref{sec:Implementation}, followed by quantitative and qualitative results in \cref{sec:quan} and \cref{sec:quality}, respectively. Ablation studies and complexity analysis are presented to validate the effectiveness of the proposed algorithm in \cref{sec:ablation} and \cref{sec:complexity}, respectively.

\subsection{Implementation details}
\label{sec:Implementation}
\subsubsection{Models.}
As mentioned in \cref{sec:preliminary}, our proposed quantization framework is evaluated on Instant-NGP \cite{torch-ngp}, a well-known NeRF variant.
All activations and parameters are quantized to facilitate integer-only inference. The bitwidth is constrained within the range of $[2,32]$, as binary quantization consistently yields nonsensical rendering results. Quantization schemes for each component are presented in \cref{tab:quan_scheme}.


\subsubsection{Training details}
For layer-wise quantization where bitwidth is learned individually for each component, feature quantization rate (FQR) \cite{CADyQ}, defined as $\text{FQR} = \frac{\sum_{i\in \mathcal{M}} B_i}{M}$, is introduced to measure the quantization performance. 
To perform training in quantization mode,  
quantization parameters are initialized using the simple PTQ method. All components are initially quantized to 8-bit except for exponential activation, which is quantized to 32-bit due to its extensive range scale (see \cref{fig:intro-b}).

As we introduce per-defined  $\mathcal{L}^\text{metric}$, the compression rate can be manipulated according to specific requirements.
Based on the average full precision loss among training set $\mathcal{L}^{\text{NeRF}}_\text{fp}$, we conduct experiments with two scenarios:

\textit{Minimal Degradation bitwidth Learning (MDL).}
For application with high-fidelity requirements, we need to maintain accuracy while minimizing bitwidth.
The metric is thus defined as 
    \begin{equation}
        \mathcal{L}^{\text{metric}}=\mathcal{L}^{\text{NeRF}}_\text{fp}.
    \end{equation}

\textit{Metric-Guided bitwidth Learning (MGL).}
As numerous studies strive for ever-higher accuracy, there inevitably arises surplus accuracy for applications with varying precision requirements, such as LOD and rendering on resource-constrained edge devices. The surplus accuracy can be traded off for efficiency by quantizing into lower bitwidth. 
In this case, the metric is defined as
    \begin{equation}
        \mathcal{L}^{\text{metric}}>\mathcal{L}^{\text{NeRF}}_\text{fp}.
    \end{equation}


\begin{table}[tb]
    \caption{A-CAQ results for metric-guided bitwidth learning on different datasets including Synthetic-NeRF \cite{NeRF}, RTMV \cite{RTMV} and Mip-NeRF360 \cite{mipnerf360}.}
    \tabcolsep=0.018\linewidth
    \label{tab:A-QAT}
    \fontsize{7}{8}\selectfont
    \centering
    \begin{tabular}{p{1cm}cccccccc}
        \toprule
        \multirow{2}{*}{\centering Dataset} 
        & \multicolumn{2}{c}{Full precision}
        & \multicolumn{2}{c}{MDL}
        & \multicolumn{2}{c}{MGL ($ 10^{-3.2}$)} 
        & \multicolumn{2}{c}{MGL ($ 10^{-3}$)}  
        \\
        \cmidrule[0.5pt](lr){2-3} \cmidrule[0.5pt](lr){4-5} \cmidrule[0.5pt](lr){6-7} \cmidrule[0.5pt](lr){8-9}
        & \centering $\text{PSNR}_\uparrow $ 
        & \centering $\text{FQR}_\downarrow$ 
        & \centering $\text{PSNR}_\uparrow$ 
        & \centering $\text{FQR}_\downarrow$ 
        & \centering $\text{PSNR}_\uparrow$ 
        & \centering $\text{FQR}_\downarrow$ 
        & \centering $\text{PSNR}_\uparrow$ 
        & {\centering $\text{FQR}_\downarrow$} \\
        \midrule
        chair     & 35.06 & 32.00 & 34.57 & 7.60 & 29.63 & 4.80 & 27.23 & 4.60
        \\
        V8        & 27.68 & 32.00 & 27.39 & 8.00 & 27.39 & 7.40 & 26.16 & 5.67
        \\
        bonsai    & 28.13 & 32.00 & 27.49 & 7.00 & 27.73 & 7.07 & 26.50 & 5.73
        \\
        \bottomrule
    \end{tabular}
\end{table}
\begin{table}[tb]
  \caption{
  \textbf{Quantitative comparisons.} 
  Instant-NGP quantized with PTQ\cite{ICARUS}, LSQ+\cite{LSQ+, MCUNeRF, TinyNeRF}, and A-CAQ are compared.
  The FQR and PSNR are reported to measure the complexity and accuracy, respectively.
  The results demonstrate that proposed methods succeed in reducing FQR while minimizing accuracy degradation.
  }
  \label{tab:comparsion}
  \fontsize{7}{8}\selectfont
  \centering
  \setlength{\tabcolsep}{1.7mm}
  \begin{tabular}{cp{2.6cm}cccccc}
  
    \toprule
    \multicolumn{2}{c}{\multirow{2}{*}{\centering Method} } 
    & \multicolumn{2}{c}{Synthetic-NeRF}
    & \multicolumn{2}{c}{Mip-NeRF360} 
    & \multicolumn{2}{c}{RTMV (V8)}  
    \\ 
    \cmidrule[0.5pt](lr){3-4} \cmidrule[0.5pt](lr){5-6} \cmidrule[0.5pt](lr){7-8} 
    \multicolumn{2}{c}{\ }
    & \centering $\text{FQR}_\downarrow$ 
    & \centering $\text{PSNR}_\uparrow$ 
    & \centering $\text{FQR}_\downarrow$ 
    & \centering $\text{PSNR}_\uparrow$ 
    & \centering $\text{FQR}_\downarrow$ 
    & {\centering $\text{PSNR}_\uparrow$}  
    \\
    \midrule
    & NGP       & 32.00 & 32.42 & 32.00 & 25.55 & 32.00 & 27.68
    \\
    \midrule
    \multirow{3}{*}{MDL} 
    & NGP-PTQ \cite{ICARUS}   & 9.60 & 31.98 & 9.60 & 25.38 & 9.60 & 27.29
    \\
    & NGP-LSQ+ \cite{LSQ+, MCUNeRF, TinyNeRF} & 9.60 & \cellcolor{green}32.11 & 9.60 & \cellcolor{green}25.48 & 9.60 & \cellcolor{green}27.40
    \\
    & NGP-A-CAQ (Ours) & \cellcolor{green}7.76 & 32.00 & \cellcolor{green}7.11 & 25.30 & \cellcolor{green}8.00 & 27.39
    \\
    \midrule
    \multirow{3}{*}{MGL\ ($ 10^{-3.2}$)}
    & NGP-PTQ   & 6.60 & 22.29 & 7.60 & 22.62 & 7.60 & 22.96
    \\
    & NGP-LSQ+  & 6.60 & 25.06 & 7.60 & 23.84 & 7.60 & 24.50
    \\
    & NGP-A-CAQ (Ours) & \cellcolor{green}5.33 & \cellcolor{green}27.58 & \cellcolor{green}6.86 & \cellcolor{green}25.18 & \cellcolor{green}7.40 & \cellcolor{green}27.39
    \\
    \midrule
    \multirow{3}{*}{MGL\ ($ 10^{-3}$)}
    & NGP-PTQ   & 5.60 & 17.75 & 6.60 & 17.54 & 5.60 & 10.54
    \\
    & NGP-LSQ+  & 5.60 & 20.26 & 6.60 & 22.44 & 5.60 & 14.36
    \\
    & NGP-A-CAQ (Ours)& \cellcolor{green}4.74 & \cellcolor{green}25.96 & \cellcolor{green}6.58 & \cellcolor{green}24.74 & \cellcolor{green}5.67 & \cellcolor{green}26.16
    \\
    \bottomrule
  \end{tabular}
\end{table}

\subsection{Quantitative Results}
\label{sec:quan}
In this section, we evaluate the performance of our A-CAQ algorithm on different datasets, including Synthetic-NeRF \cite{NeRF}, RTMV \cite{RTMV}, and Mip-NeRF360 \cite{mipnerf360}. These datasets comprise scenes with varying levels of complexity.


The experiments begin with various accuracy requirements to demonstrate the capability and flexibility of the metric-guided feature. The experimental results are presented in \cref{tab:A-QAT}. Quantization schemes of different accuracy can be effectively learned with different $\mathcal{L}^\text{metric}$. MSE loss is selected as a general metric here. Note that other quality metrics such as PSNR, Structural Similarity Index (SSIM \cite{SSIM}) and Learned Perceptual Image Path Similarity (LPIPS \cite{LPIPS}) can also be used according to the specific requirements of different applications.



To validate the effectiveness of our A-CAQ in consistently producing efficient quantization schemes while maintaining requisite quality, we compare it with PTQ and LSQ+ schemes with different bitwidth configurations, which are widely used for radiance field model quantization \cite{ICARUS, MCUNeRF, TinyNeRF}.
As shown in \cref{tab:comparsion}, the proposed methods can identify optimal quantization schemes for various scenarios. For high-fidelity applications (labeled as MDL), integer-only rendering is achieved with negligible PSNR loss ($<0.5$ dB), while the FQR is significantly reduced. In resource-constrained scenarios (labeled as MGL), our method achieves state-of-the-art results for both accuracy and efficiency.



\begin{figure}[tb]
  \centering
  \includegraphics[width=1.0\textwidth]{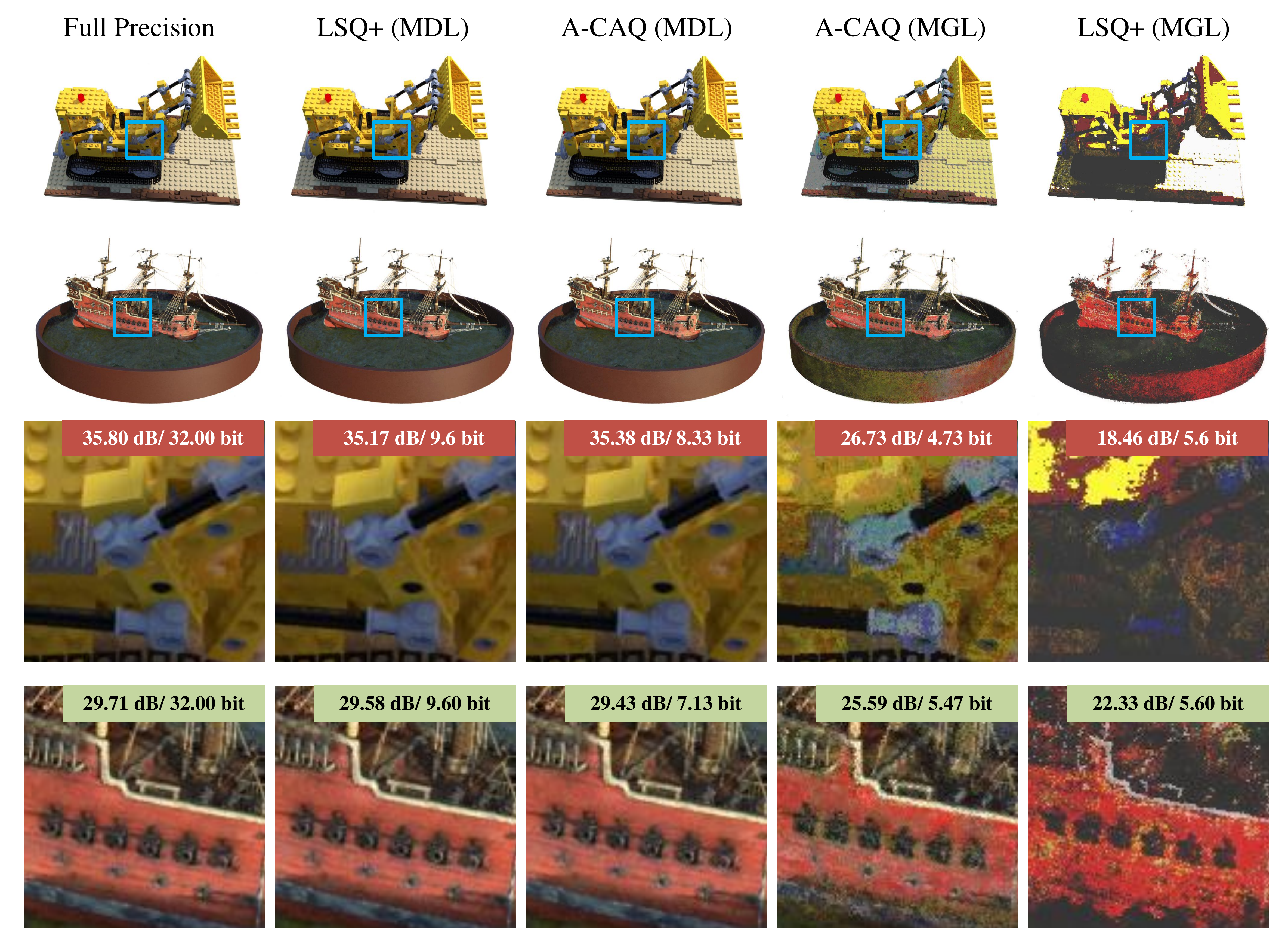}
  \caption{
  \textbf{Qualitative results} of the "lego" and "ship" datasets from Synthetic-NeRF.
  Proposed A-CAQ outputs visual clean results for both MDL and MGL scenarios.
  }
  \label{fig:quality}
\end{figure}

\begin{figure}[tb]
  \centering
  \includegraphics[width=1.0\textwidth]{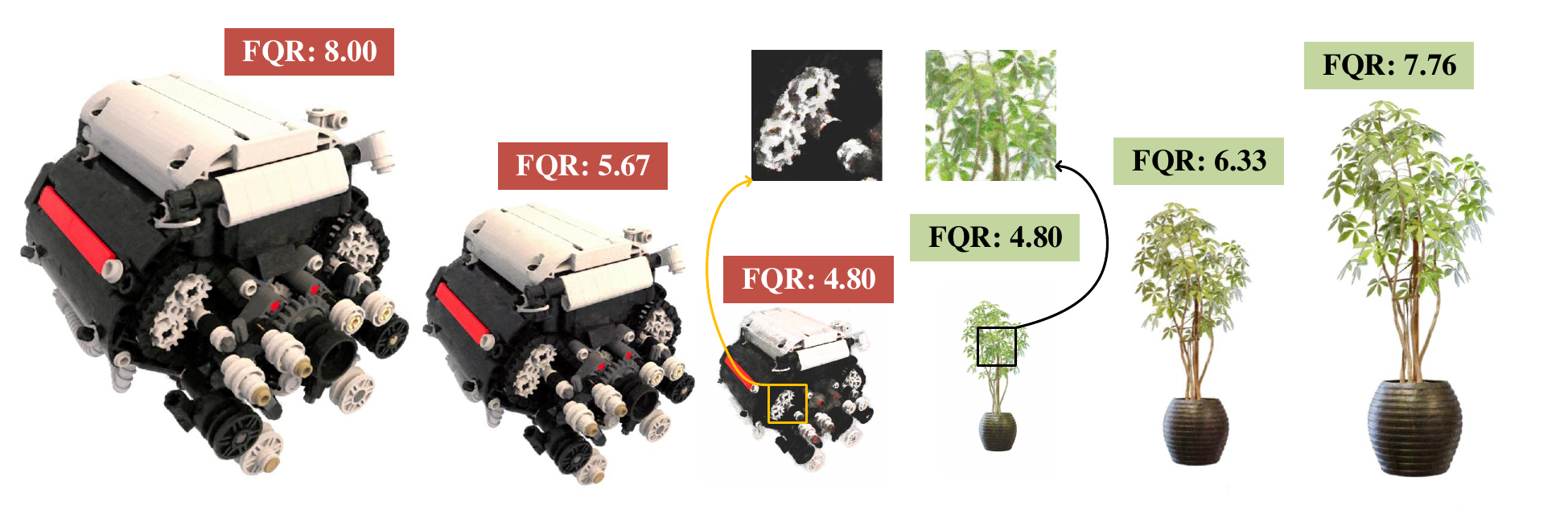}
  \caption{
  \textbf{Quantized LOD.}
  Our proposed method can compress scenes in LOD style. The detail levels can be easily manipulated considering available resources and required accuracy.
  }
  \label{fig:LOD}
\end{figure}

\begin{figure}[tb]
    \centering
    \begin{minipage}{\linewidth}
        \centerline{\includegraphics[width=0.9\textwidth] 
        {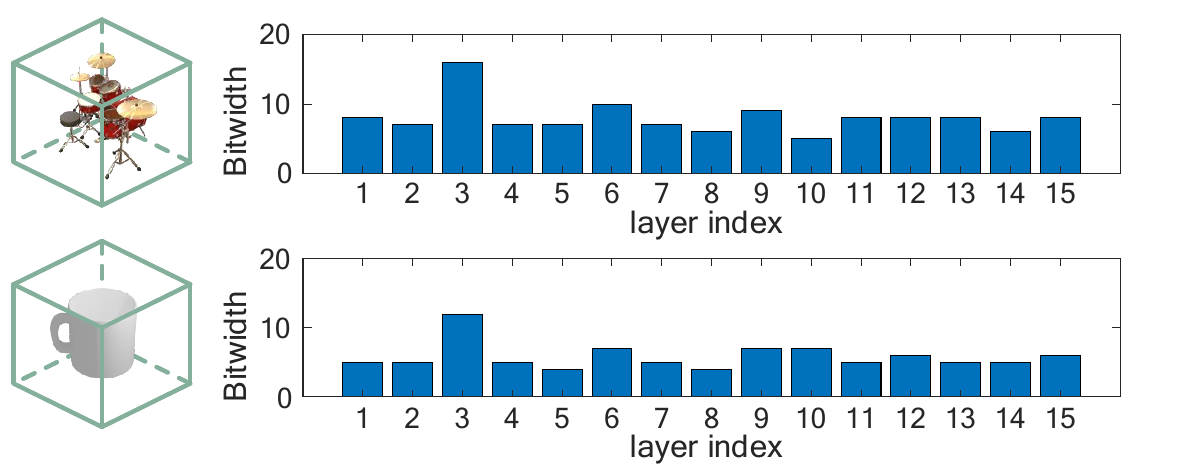} }
    \end{minipage}
    \caption{
    \textbf{Visualization of content-aware layer-wise quantization.}
    The total number of components in our radiance field pipeline $M=15$ including codebook and MLPs.
  }
  \label{fig:layer_wise}
\end{figure}

\subsection{Qualitative Results}
\label{sec:quality}

We also present qualitative results and comparisons with other widely used methods. As depicted in \cref{fig:quality}, our A-CAQ consistently yields visually clean results across various loss metrics while requiring fewer bits. In contrast, existing quantization paradigms introduce significant distortion with comparable bitwidth configurations, as they are unable to discern quantization sensitivities among different scenes and layers.


To demonstrate the effectiveness and flexibility of our method, LOD rendering results are presented in \cref{fig:LOD}. Our dynamic quantization scheme can simultaneously filter and compress radiance field models, proving advantageous for LOD tasks \cite{VQ-AD}. Furthermore, quantization results for scenes with varying complexities are illustrated in \cref{fig:layer_wise}. Content complexity influences the quantization sensitivity for output features of each layer, which can be distinguished with the proposed A-CAQ algorithm.

\subsection{Ablation study}
\label{sec:ablation}
\textbf{Scene-wise and layer-wise quantization.}
To verify the significance of scene-wise and layer-wise quantization, we compare our proposed quantization framework with fixed-bitwidth QAT schemes, whose bitwidths are determined through a trial and error approach. Results are presented in \cref{tab:ablation_layer}. Quantizing all scenes with a fixed bitwidth throughout the entire pipeline results in performance loss and resource wastage (model \textbf{(1a)}). To accommodate the varying complexities observed in different scenes, a manual adjustment of scene-specific bitwidth is conducted, which mitigates the degradation in quantization (model \textbf{(1b)}). Utilizing A-CAQ, scene-dependent layer-wise quantization bitwidths are effectively learned. Following the MDL approach, the lowest average bitwidth schemes are learned while maintaining accuracy (model \textbf{(Ours, 1e)}). As a comparison, we quantize all scenes with the same conservative scheme: quantizing each layer with the highest layer-wise bitwidth learned by A-CAQ among all scenes (model \textbf{(Ours, 1d)}). This achieves the highest PSNR among all models while sacrificing extra model complexity.


\subsubsection{Adversarial learning.}
Our A-CAQ presents a multi-task learning problem.
Here, we analyze the effects of two optimization tasks individually, which are reported in \cref{tab:ablation_loss}. Quantization without any post-processing results in the worst accuracy drop with the highest computational consumption (model \textbf{(2a)}). Post-training can, in turn, alleviate the performance degradation (model \textbf{(2b)}). Minimizing $\mathcal{L}^\text{bit}$ allows for learning the layer-wise quantization bitwidth, which searches for the lowest bitwidth in the PTQ performance space (model \textbf{(2c)}). Our proposed method can achieve much lower bitwidths according to the required accuracy as we search for bitwidth in the QAT performance space.


\begin{table*}
    \begin{floatrow}
        \capbtabbox{
            \fontsize{6}{11}\selectfont
            \begin{tabular}{p{1.5cm}|cc|cc}
                \toprule
                { } 
                & layer
                & scene
                & $\text{FQR}_\downarrow$ 
                & $\text{PSNR}_\uparrow$ 
                \\
                \midrule
                \centering{\textbf{(1a)}} 
                & \color{red}{\XSolidBrush} 
                & \color{red}{\XSolidBrush} 
                & 14.00 & 31.56
                \\
                \centering{\textbf{(1b)}} 
                & \color{red}{\XSolidBrush}  
                & \color{brown}{\Checkmark} 
                & 14.27 & 32.04
                \\
                \centering{\textbf{(LSQ+, 1c)}} 
                & \color{brown}{\Checkmark} 
                & \color{red}{\XSolidBrush}  
                & 9.60 & 32.11
                \\
                \centering{\textbf{(Ours, 1d)}} 
                & \color{brown}{\Checkmark} 
                & \color{red}{\XSolidBrush}  
                & 8.93 & \cellcolor{green}32.30
                \\
                \centering{\textbf{(Ours, 1e)}} 
                & \color{brown}{\Checkmark} 
                & \color{brown}{\Checkmark} 
                & \cellcolor{green}7.76 & 32.00
                \\
                \bottomrule
            \end{tabular}
        }{
            \caption{Ablation study on layer-wise and scene-wise quantization.}
            \label{tab:ablation_layer}
        }
        \capbtabbox{
            \fontsize{6}{10}\selectfont
            \begin{tabular}{p{1cm}|cccccc}
                \toprule
                \multirow{2}{*}{ } 
                & \multicolumn{2}{c}{loss}
                & \multicolumn{2}{c}{MDL}
                & \multicolumn{2}{c}{MGL}
                \\
                \cmidrule[0.5pt](lr){2-3}
                \cmidrule[0.5pt](lr){4-5}
                \cmidrule[0.5pt](lr){6-7}
                & $\mathcal{L}^{\text{NeRF}}$ 
                & $\mathcal{L}^{\text{bit}}$ 
                & $\text{FQR}_\downarrow$
                & $\text{PSNR}_\uparrow$
                & $\text{FQR}_\downarrow$
                & $\text{PSNR}_\uparrow$
                \\
                \midrule
                \centering{\textbf{(2a)}} 
                & \color{red}{\XSolidBrush}
                & \color{red}{\XSolidBrush}
                & 9.60 & 31.98 & 6.60 & 22.29
                \\
                \centering{\textbf{(2b)}} 
                & \color{brown}{\Checkmark} 
                & \color{red}{\XSolidBrush}
                & 9.60 & 32.11 & 6.60 & 25.06
                \\
                \centering{\textbf{(2c)}} 
                & \color{red}{\XSolidBrush} 
                & \color{brown}{\Checkmark} 
                & 8.07 & 32.00 & 6.43 & 27.95
                \\
                \centering{\textbf{Ours}} 
                & \color{brown}{\Checkmark} 
                & \color{brown}{\Checkmark} 
                & \cellcolor{green}7.76 & 32.00 & \cellcolor{green}5.33 & 27.58
                \\
                \bottomrule
              \end{tabular}
        }{
          \caption{Ablation study on the adversarial losses: $\mathcal{L}^{\text{NeRF}}$ and $\mathcal{L}^{\text{bit}}$.}
          \label{tab:ablation_loss}
        }
    \end{floatrow}
\end{table*}

\subsection{Complexity analysis}
\label{sec:complexity}

Our proposed quantization method can determine the optimal content-dependent bitwidth for both high-fidelity (MDL) and resource-constrained (MGL) scenarios. To demonstrate how bitwidth influences resource overhead, 
we use the number of operations, weighted by bitwidth (BitOps) \cite{BitOps}, involved in multiplications as a metric for computational complexity
which has been widely employed for mixed-precision network systems \cite{BitOps_case1, BitOps_case2, BitOps_case3}. Additionally, model sizes are listed to reflect memory consumption.

The results presented in \cref{tab:BitOps} demonstrate the efficiency of our method.
In the MDL scenario, our framework achieves a reduction of $\sim90.78\%$ in BitOps compared to the baseline, and $\sim11.66\%$ reduction compared to quantization with LSQ+. Regarding memory consumption, our proposed A-CAQ achieves a reduction of $\sim77.68\%$ compared to the baseline, and $\sim10.74\%$ reduction compared to LSQ+.
In the MGL scenario, the reduction expands to $\sim94.45\%$ and $\sim16.17\%$ in BitOps compared to NGP and NGP-LSQ+ respectively. Moreover, there is a reduction of $\sim89.43\%$ and $\sim 32.34\%$ in storage compared to NGP and NGP-LSQ+ respectively. Notably, our method also achieves a 0.9 dB increase in PSNR compared to NGP-LSQ+.



\begin{table}[tb]
  \caption{
  \textbf{Complexity analysis of A-CAQ for inference.}
  BitOps \cite{BitOps} for rendering one 800 $\times$ 800 image and model storage are measured for time and space consumption, respectively.
  }
  \label{tab:BitOps}
  \fontsize{7}{8}\selectfont
  \setlength{\tabcolsep}{1.1 mm}
  \centering
  \begin{tabular}{p{2.6cm}|cccccc}
    \toprule
    {}
    & \multicolumn{3}{c}{MDL}
    & \multicolumn{3}{c}{MGL}
    \\
    \cmidrule[0.5pt](lr){2-4} \cmidrule[0.5pt](lr){5-7}
    { } 
    & {\centering{$\text{PSNR}_\uparrow$\ }}
    & {\centering{$\text{BitOps}_\downarrow$\ [\text{T}]} }
    & {\centering{$\text{Storage}_\downarrow$\ [\text{MB}]} }
    & {\centering{$\text{PSNR}_\uparrow$\ }}
    & {\centering{$\text{BitOps}_\downarrow$\   }}
    & {\centering{$\text{Storage}_\downarrow$\  }}
    \\
    \midrule
    NGP \cite{NGP}
    & \cellcolor{green}32.42
    & 71.01
    & 46.56
    & -
    & -
    & - 
    \\
    NGP-LSQ+ \cite{LSQ+, MCUNeRF, TinyNeRF}
    & 32.11
    & 7.41
    & 11.64
    & 25.06
    & 4.70
    & 7.28
    \\
    NGP-A-CAQ (Ours)
    & 32.00
    & \cellcolor{green}6.55
    & \cellcolor{green}10.39
    & \cellcolor{green}25.96
    & \cellcolor{green}3.94
    & \cellcolor{green}4.92
    \\
    \bottomrule
  \end{tabular}
\end{table}

\section{Conclusion}
In this work, we introduce the concept of content-aware radiance field and explore the relationship between 3D scene complexity and quantization schemes. This motivates us to propose the A-CAQ algorithm, which learns bitwidth using gradients obtained by automatically perceiving the scene.  
Our method dynamically allocates layer-wise and scene-wise bitwidths.
Experimental results demonstrate that the proposed algorithm significantly reduces model complexity for various scenarios through quantization, under different requirements.

\subsubsection{Limitations.}
This study explores the novel concept of content-aware radiance fields, deftly integrating mixed-precision quantization into its framework. Despite the progress made, the realm of content-aware radiance fields beckons with unexplored territories ripe for detailed examination.
For instance, by conducting a targeted search of network architecture aligned with reliable indicators that reflect the complexity of 3D scenes, it becomes feasible to construct a more comprehensive content-aware radiance field framework.


\section*{Acknowledgements}
This work was supported by the Central Guided Local Science and Technology Foundation of China (YDZX20223100001001).

%
%
\bibliographystyle{splncs04}
\bibliography{egbib}

\appendix

\section{Additional details of LBQ and A-CAQ}
\label{sec:gradient}
\subsection{Gradients for differentiable quantization}
We introduce three quantization schemes for three different components within the radiance field pipeline in our proposed LBQ framework.
To efficiently back-propagate through the simulated quantizer block, derivatives to three sets of key parameters are derived as follows
\subsubsection{Neural weights.}
The weights are quantized with signed symmetric quantization, and the partial derivatives to the key parameters $r_v$ and $b$ are given as
\begin{equation}
    \label{eq:grad_rv}
    \frac{\partial \hat{v}}
    {\partial r_v} = 
    \left\{
        \begin{aligned}
        &\frac{1}{r_v} \left( s \cdot \left\lfloor v/s \right\rceil - v \right)
        \quad v_\text{min}\leq v\leq v_\text{max}\\
        &\frac{q_\text{max}}{r_q} \quad v > v_\text{max} \\
        &\frac{q_\text{min}}{r_q} \quad v < v_\text{min}
        \end{aligned}
    \right. ,
\end{equation}
\begin{equation}
    \label{eq:grad_b}
    \frac{\partial \hat{v}}
    {\partial b} = 
    \left\{
        \begin{aligned}
        &\frac{2^B\ln2}{r_q} \left(v- s \cdot \left\lfloor v/s \right\rceil\right)  \quad v_\text{min}\leq v\leq v_\text{max}
        \\
        &\frac{2^B\ln2}{r_q} \left( v_\text{max} - q_\text{max} \cdot s \right) \quad v > v_\text{max} 
        \\
        &\frac{2^B\ln2}{r_q} \left( v_\text{min} - q_\text{min} \cdot s \right) 
        \quad v < v_\text{min}
        \end{aligned}
    \right. .
\end{equation}

\subsubsection{ReLU and exponential activations.}
The ReLU and exponential activations are quantized with unsigned symmetric quantization, the partial derivatives are derived as 
\begin{equation}
    \label{eq:grad_rv}
    \frac{\partial \hat{v}}
    {\partial r_v} = 
    \left\{
        \begin{aligned}
        &\frac{1}{r_v} \left( s \cdot \left\lfloor v/s \right\rceil - v \right)
        \quad v_\text{min}\leq v\leq v_\text{max}\\
        &1 \quad v > v_\text{max} \\
        &0 \quad v < v_\text{min}
        \end{aligned}
    \right. ,
\end{equation}
\begin{equation}
    \label{eq:grad_b}
    \frac{\partial \hat{v}}
    {\partial b} = 
    \left\{
        \begin{aligned}
        &\frac{2^B\ln2}{r_q} \left(v- s \cdot \left\lfloor v/s \right\rceil\right)  
        \quad v_\text{min}\leq v\leq v_\text{max}
        \\
        & 0
        \quad \text{otherwise} 
        \end{aligned}
    \right. .
\end{equation}

\subsubsection{PE and others.}
PE and other components are quantized with asymmetric quantization which is regarded as a general form.
Considering the offset is equivalently replace by $v_\text{max}$, the derivatives \wrt key parameters are given as
\begin{equation}
    \label{eq:grad_vmax}
    \frac{\partial \hat{v}}
    {\partial v_\text{max}} = 
    \left\{
        \begin{aligned}
        &0 \quad v_\text{min}\leq v\leq v_\text{max}\\
        &1 \quad \text{otherwise} 
        \end{aligned}
    \right. ,
\end{equation}
\begin{equation}
    \label{eq:grad_rv}
    \frac{\partial \hat{v}}
    {\partial r_v} = 
    \left\{
        \begin{aligned}
        &\frac{1}{r_v} \left( s \cdot \left\lfloor v/s \right\rceil - v \right)
        \quad v_\text{min}\leq v\leq v_\text{max}\\
        &1-\frac{v_\text{max}}{r_v}-\frac{z}{r_q} \quad v > v_\text{max} \\
        &-\frac{v_\text{max}}{r_v}-\frac{z}{r_q} \quad v < v_\text{min}
        \end{aligned}
    \right. ,
\end{equation}
\begin{equation}
    \label{eq:grad_b}
    \frac{\partial \hat{v}}
    {\partial b} = 
    \left\{
        \begin{aligned}
        &\frac{2^B\ln2}{r_q} \left(v- s \cdot \left\lfloor v/s \right\rceil\right)  \quad v_\text{min}\leq v\leq v_\text{max}
        \\
        &\frac{2^B\ln2}{r_q} \left(v_\text{min} + s\cdot z\right) 
        \quad \text{otherwise} 
        \end{aligned}
    \right. .
\end{equation}

\subsection{Pseudo code for A-CAQ}
Our proposed A-CAQ is a multi-task learning-based algorithm, and the pseudo codes are shown in \cref{alg:A-CAQ}.

\begin{algorithm}[hbt]  
  \caption{A-CAQ}  
  \label{alg:A-CAQ}  
  \begin{algorithmic}  
    \Require  
        Trained model with initial quantization parameters $\mathcal{Q}^{0,0}=\{\mathbf{\Omega}^{0,0}, \mathbf{v}_\text{max}^{0,0}, \mathbf{r}_v^{0,0} \}$;
        initial soft bitwidth $\mathbf{b}^{0,0}$;
        training set $\mathcal{D}$; learning rate $\beta$;
        hyper-defined metric $\mathcal{L}^\text{metric}$
    \Ensure The convergent model parameters $\mathcal{Q}^{E-1,I}, \mathbf{b}^{E-1,I}$

    \For{epoch $i = 0$ to $E-1$}
        \For{iteration $j=1$ to $I$}
            \State $\mathcal{B}_j \gets \mathcal{D}$ 
            \Comment{Get Batch using ray-marching}
            \State $(\sigma, \mathbf{c}) \gets 
            F(\mathbf{x}, \mathbf{d}\mid \mathcal{Q}^{i,j-1}, b^{i,j-1})$ for all $(\mathbf{x}, \mathbf{d}) \in \mathcal{B}_j $
            \State $\hat{C}(\ell) \gets$ volume rendering for all $\ell \in \mathcal{R}$
            
            \State ${\nabla}_j^{(1)} \gets {\nabla}_{\mathcal{Q}^{i,j-1}}\mathcal{L}^\text{NeRF}(\hat{C}(\ell)\mid\mathcal{Q}^{i,j-1}, b^{i,j-1})$
            
            \State ${\nabla}_j^{(2)} \gets {\nabla}_{b^{i,j-1}}\mathcal{L}^\text{bit}(\hat{C}(\ell)\mid\mathcal{Q}^{i,j-1},b^{i,j-1})$

            \State $\mathcal{Q}^{i,j} \gets \mathcal{Q}^{i,j-1} - \frac{\beta}{\lvert \mathcal{B}_j \rvert} {\nabla}_j^{(1)}$
            \State $\mathbf{b}^{i,j} \gets \mathbf{b}^{i,j-1} - \frac{\beta}{\lvert \mathcal{B}_j \rvert} {\nabla}_j^{(2)}$
            \Comment{Update with gradients averaged on batch size}
        \EndFor
  	\EndFor
  \end{algorithmic}  
\end{algorithm}

\section{More Implementation Details}
Given that our framework is optimized based on a well-trained full-precision model, the learning rates for the two tasks are differently configured.
Specifically, 
as the bitwidths are trained from scratch, the initial bitwidth learning rates are set to $1\times 10^{-2}$, while those for other parameters are $1\times10^{-3}$.
The summation of bitwidth penalties is fixed to $\sum_i \epsilon_i = 1\times 10^{-3}$. And training of A-CAQ finishes after 3000 iterations.
Other training configurations, such as batch size,  learning rate schedule and optimizer, follow the settings of the baseline model.

\section{Details on penalty}
\begin{figure}[htb]
    \centering
    \begin{subfigure}{0.32\linewidth}
        \centerline{\includegraphics[width=\textwidth] 
            {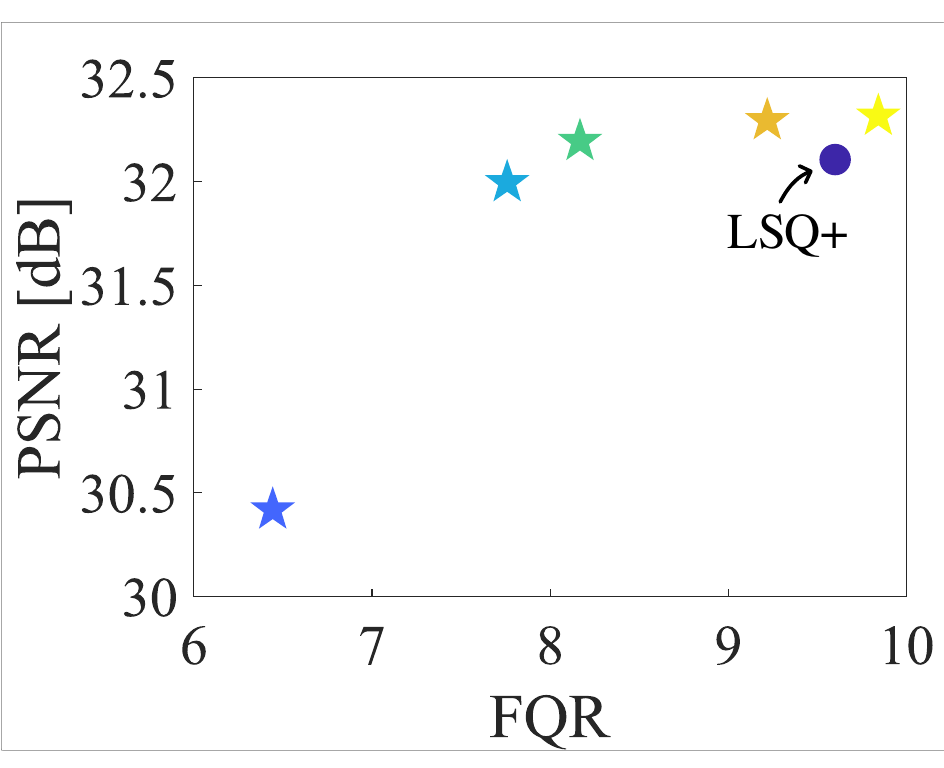} }
        \caption{MDL.}
        \label{fig:penalty-a}
    \end{subfigure}
    \hfill
    \begin{subfigure}{0.32\linewidth}
        \centerline{\includegraphics[width=\textwidth] 
            {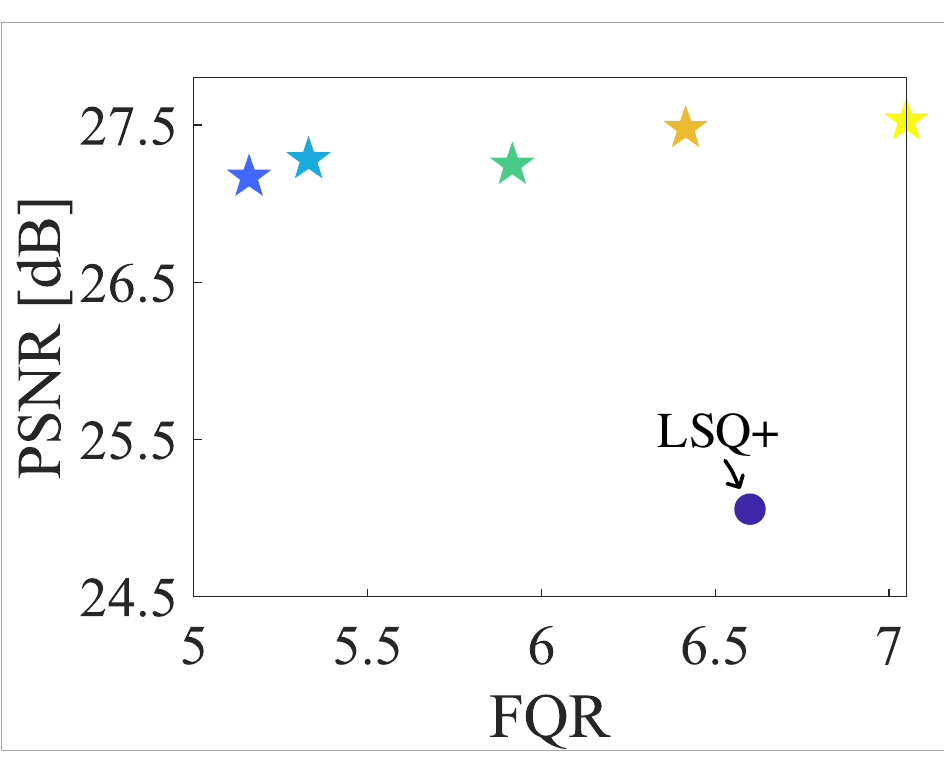} }
        \caption{MGL ($10^{-3.2}$).}
        \label{fig:penalty-b}
    \end{subfigure}
    \hfill
    \begin{subfigure}{0.32\linewidth}
        \centerline{\includegraphics[width=\textwidth] 
            {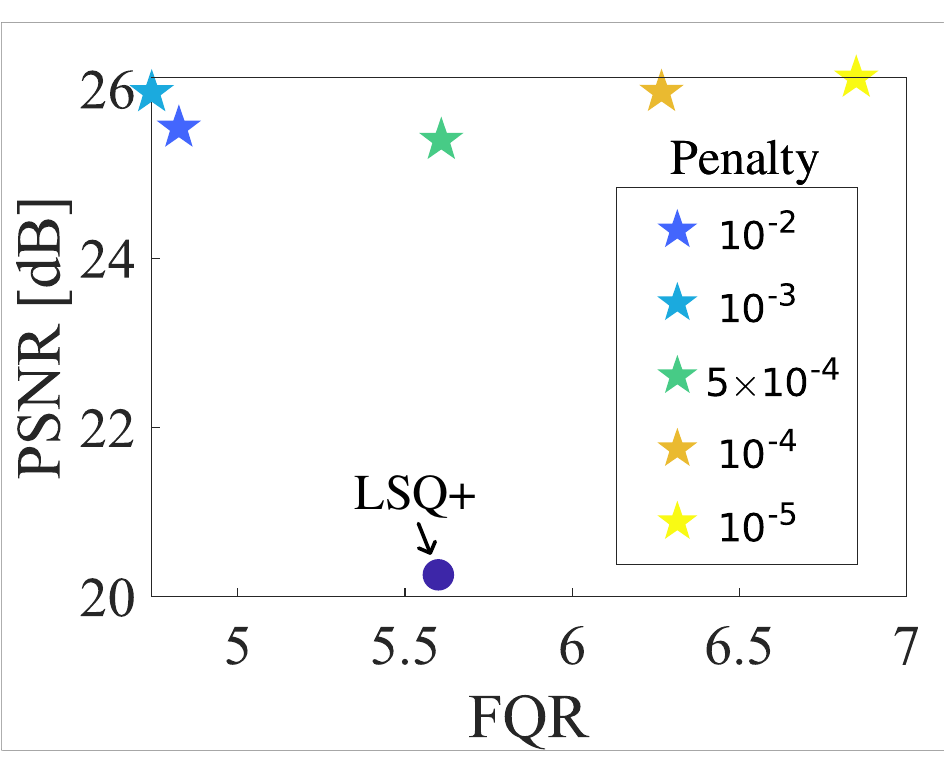} }
        \caption{MGL ($10^{-3}$).}
        \label{fig:penalty-c}
    \end{subfigure}
    \caption{
    Penalty effects on A-CAQ.
  }
  \label{fig:penalty}
\end{figure}

The summation of bitwidth penalties serves as a per-defined parameters governs the balance between the accuracy and efficiency.
To illustrate its impact, experimental results on Synthetic-NeRF dataset are shown in \cref{fig:penalty}.
Lower penalties result in higher accuracy, albeit at the expense of efficiency. 
Moreover, our proposed A-CAQ outperforms LSQ+ in terms of both accuracy and efficiency, offering a straightforward means to achieve higher levels of both metrics.

\clearpage
\section{More qualitative results}
\begin{figure}[htb]
  \centering
  \includegraphics[width=1.0\textwidth]{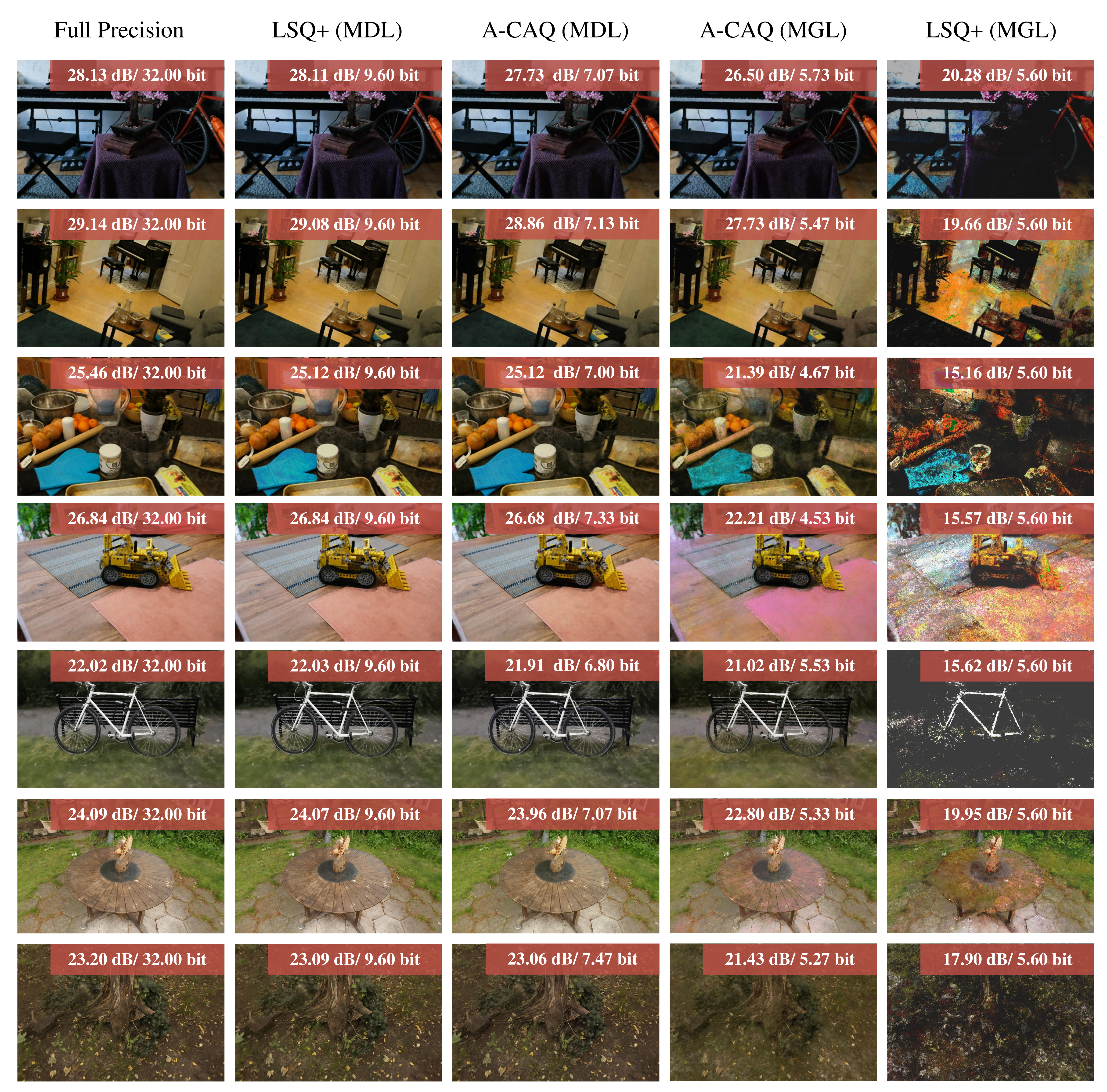}
  \caption{
  Rendering results on \textbf{Mip-NeRF360} dataset.
  }
  \label{fig:mip-nerf360}
\end{figure}

\begin{figure}[tb]
  \centering
  \includegraphics[width=1.0\textwidth]{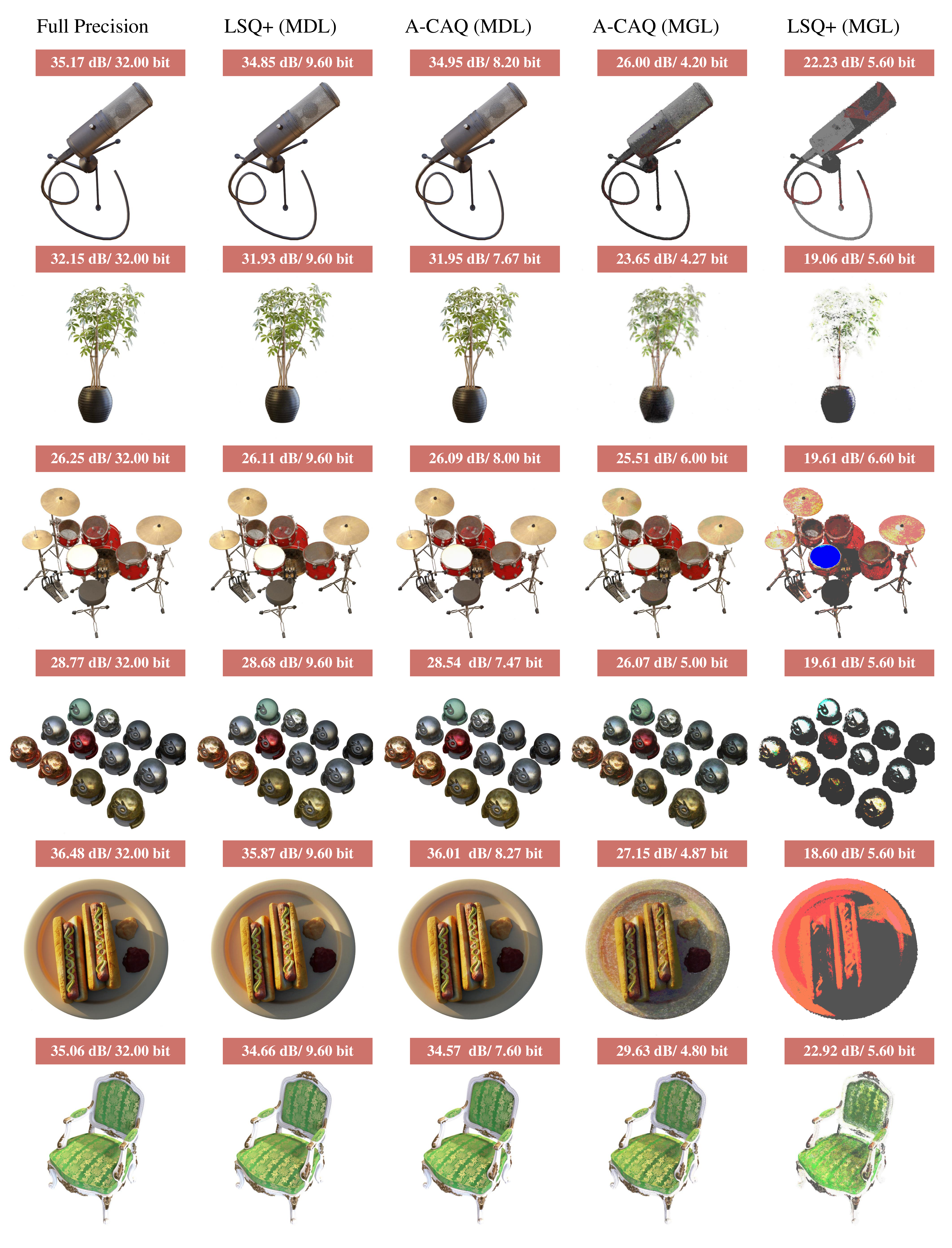}
  \caption{
  Rendering results on \textbf{Synthetic-NeRF} dataset.
  }
  \label{fig:synthetic-nerf}
\end{figure}

\cref{fig:mip-nerf360} and \cref{fig:synthetic-nerf} provide additional qualitative results in Mip-NeRF360 and Synthetic-MeRF dataset.

\end{document}